\documentclass[conference]{IEEEtran}
\IEEEoverridecommandlockouts
\usepackage{cite}
\usepackage{amsmath,amssymb,amsfonts}
\usepackage{algorithmic}
\usepackage{graphicx}
\usepackage{textcomp}
\usepackage{xcolor}
\usepackage{xspace}
\usepackage{booktabs}
\usepackage{multirow}
\usepackage[strings]{underscore}
\newcommand{\ie}{\textit{i.e., }}
\newcommand{\eg}{\textit{e.g., }}
\newcommand{\model}{{EiCI-Net}\xspace}

\def\BibTeX{{\rm B\kern-.05em{\sc i\kern-.025em b}\kern-.08em
    T\kern-.1667em\lower.7ex\hbox{E}\kern-.125emX}}
\begin{document}

\title{Joint Explicit and Implicit Cross-Modal Interaction Network for Anterior Chamber Inflammation Diagnosis}

\author{
\IEEEauthorblockN{
Qian Shao\IEEEauthorrefmark{1},
Ye Dai\IEEEauthorrefmark{2},
Haochao Ying\IEEEauthorrefmark{3},
Kan Xu\IEEEauthorrefmark{2},
Jinhong Wang\IEEEauthorrefmark{1},
Wei Chi\IEEEauthorrefmark{2}, and
Jian Wu\IEEEauthorrefmark{3}}

\IEEEauthorblockA{\IEEEauthorrefmark{1}College of Computer Science \& Technology, Zhejiang University, Hangzhou, China}
\IEEEauthorblockA{\IEEEauthorrefmark{2}State Key Laboratory of Ophthalmology, Sun Yat-sen University}
\IEEEauthorblockA{\IEEEauthorrefmark{3}Second Affiliated Hospital School of Medicine, School of Public Health, Zhejiang University, Hangzhou, and\\
Institute of Wenzhou, Zhejiang University, Wenzhou, China}
\IEEEauthorblockA{Qian Shao and Ye Dai contributed equally.}
\IEEEauthorblockA{Corresponding Authors: Haochao Ying, Wei Chi \quad Emails: haochaoying@zju.edu.cn, chiwei@mail.sysu.edu.cn}
}

\maketitle

\begin{abstract}
Uveitis demands the precise diagnosis of anterior chamber inflammation (ACI) for optimal treatment. Current diagnostic methods only rely on single-modal data, which leads to poor performance. This may be attributed to the current lack of publicly available multi-modal datasets and the lack of a fusion method for the task of diagnosis for ACI. Specifically, existing fusion methods focus on empowering implicit interactions (\ie self-attention and its variants) but neglect explicit cross-modal interactions, especially clinical knowledge and imaging properties, which are crucial to the diagnosis task. To this end, we propose a joint \underline{E}xplicit and \underline{i}mplicit \underline{C}ross-Modal \underline{I}nteraction \underline{Net}work (EiCI-Net), which uses anterior segment optical coherence tomography (AS-OCT) images, slit-lamp images, and clinical indicators (tabular data) as input. In detail, we develop CNN-Based Encoders and a Tabular Processing Module (TPM) to extract feature representations of different modalities. Next, we devise an Explicit Cross-Modal Interaction Module (ECIM) to generate attention maps as a kind of explicit clinical knowledge based on the tabular feature maps, then integrate them into the slit-lamp feature maps, allowing the model to focus on more effective informativeness of the slit-lamp images. After that, the Implicit Cross-Modal Interaction Module (ICIM), a transformer-based network, implicitly enhances modality interactions. Finally, we construct a considerable real-world multimodal dataset, which will be made available publicly, and conduct extensive experiments to evaluate the superior performance of EiCI-Net compared with other methods. In addition, the experimental results demonstrate the effectiveness of explicit cross-modal interaction, which has always been ignored before.
\end{abstract}


\section{Introduction}
Uveitis, the inflammation inside the eye, accounts for approximately $10\%$ of blindness worldwide, which occurs in every age group, including even children~\cite{epidemiology, prevalence, acute}. 
Uveitis can be anatomically and easily classified into anterior, intermediate, posterior, and panuveitis by the primary site of inflammation. The accurate diagnosis of inflammation type can effectively guide therapy.
Prior work demonstrates that community ophthalmologists overwhelmingly see anterior uveitis ($91\%$) in clinical~\cite{causes2}.
Thus, the anterior chamber inflammation (ACI) diagnosis and classification of disease severity are crucial for choosing appropriate therapeutic measures.

\begin{figure*}[t]
    \centering
    \includegraphics[width=1\textwidth]{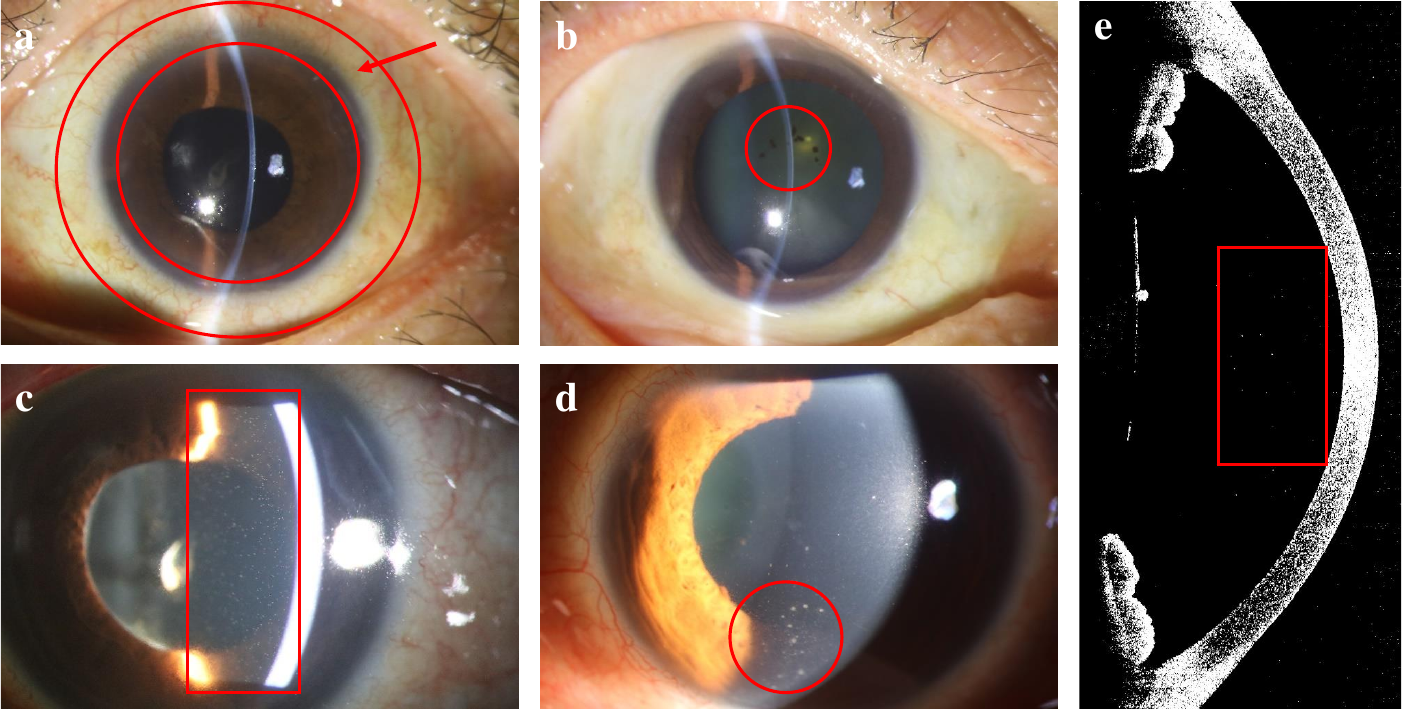}
    \caption{Comparison of slit-lamp images (a$\sim$d) and AS-OCT images (e): a. ciliary haemorrhage at the limbus; b. Iris pigmentation; c. AC cells; d. KPs; e. AC cells. Slit-lamp images, despite containing rich information indicative of ACI, challenge automated algorithms to detect AC cells due to their complex backgrounds and the small area of AC cells. Conversely, AC cells are more detectable in AS-OCT images, though without portraying the comprehensive state outside the eyeball state and iris pigmentation.}
    \label{figcon}
\end{figure*}

Recently, the Standardization of Uveitis Nomenclature (SUN) scoring system has been widely used to grade ACI by slit-lamp images in clinical~\cite{Standardization}. Slit-lamp images can reflect not only the overall state outside the eyeball but also various important indicators inside the anterior chamber (AC) for diagnosing ACI (shown in Fig.~\ref{figcon} a$\sim$d), including the number of AC cells, iris pigmentation, keratic precipitates (KPs), and so on. However, the SUN system is a subjective scoring and grading system that can easily lead to marked differences by interobservers depending on different experiences~\cite{Interobserver}. Additionally, when the severity level of ACI is high, the distribution of AC cells becomes relatively dense, which makes it both time-consuming and labour-intensive for doctors to diagnose ACI by counting these cells using the naked eye, as shown in Fig.~\ref{figcon}c. As a result, automatically diagnosing and classifying ACI needs to be developed.

However, it is very challenging for current automated algorithms to accurately detect and count these cells in slit-lamp images due to the complexity of the background in slit-lamp images and the small area occupied by AC cells within the images, as shown in Fig.~\ref{figcon}c.
Consequently, an increasing number of researchers resort to anterior segment optical coherence tomography (AS-OCT) image analysis~\cite{objective, automated, analysis}.
The background information in AS-OCT images is simple, allowing for accurate and easy segmentation of the AC. After applying threshold processing, AC cells can be detected as demonstrated in Fig.~\ref{figcon}e.
Nevertheless, AS-OCT imaging serves as a cross-sectional imaging technique for the AC, generating hundreds of AS-OCT images per eye (\ie 128 images in our experiments), while some indicators (\eg KPs) typically appear on a limited section of these cross-sectional AS-OCT images. 
In addition, AS-OCT images solely depict the internal features of the AC without the ability to reflect the overall state outside the eyeball and iris pigmentation, which also helps the classification of ACI.

In summary, slit-lamp images offer more comprehensive information to identify the presence of ACI, but quantifying detailed features using automated algorithms is challenging such as the number of AC cells. On the other hand, AS-OCT images provide limited information, yet it is easier to employ automated algorithms to quantify the number of AC cells. Therefore, efficiently exploiting and fusing each modal strength will significantly improve prediction performance.
But, existing fusion paradigms focus on empowering implicit modality interactions (\ie self-attention and its variants), but neglect to inject explicit modality interactions, especially from expert knowledge and imaging properties. For example, Zheng et al.~\cite{multisp} and Saeed et al.~\cite{tmss} adopt a transformer to fuse multimodal data implicitly for the respective tasks. 
Thus, exploring the explicit and implicit modality interactions for ACI diagnosis needs further study.

To this end, we propose a jointly explicit and implicit cross-modal interaction network called \model, which mainly consists of Feature Encoders (CNN-Based Encoders), a Tabular Processing Module (TPM), an Explicit Cross-Modal Interaction Module (ECIM) and an Implicit Cross-Modal Interaction Module (ICIM).
Specifically, two CNN-based Encoders are used to learn rich informativeness from two types of slit-lamp images (\ie the overall state images outside the eyeball and the observation images of AC) respectively. Meanwhile, the TPM first quantifies the features of AS-OCT images which are difficult to obtain from slit-lamp images and then encodes these features together with other clinical data that may be closely related to ACI to establish tabular features as the explicit cross-modal interaction features.
Then, the ECIM generates attention maps by calculating the correlation between slit-lamp image feature maps and tabular feature maps, guiding the slit-lamp feature maps to focus more on effective task-related informativeness. 
Finally, the ICIM adopts a transformer-based network for implicit modality interactions.

The main contributions can be summarized as follows:
\begin{itemize}
    \item We propose \model to diagnose ACI using multimodal data. To our knowledge, we are the first to diagnose ACI under the fusion of multimodal data.
    \item We focus on the implicit and explicit modality interactions jointly for diagnosing ACI. The experiment demonstrates that explicit modality interactions significantly improve performance, which is always neglected.
    \item We construct the first real-world multimodal ACI diagnosis dataset for research and conduct extensive experiments to evaluate the effectiveness of \model. The results validate the superior performance over current methods.
\end{itemize}

\section{Related Work}

\subsection{Anterior Chamber Inflammation Diagnosis}
The current standard for measuring ACI is established by the SUN International Workshop, which assigns the grade of inflammation a score between 0 (no inflammation) to 4+ (severe inflammation) based on the number of visible cells in a 1 × 1 mm slit-lamp beam in the AC through the central cornea~\cite{gaci}. Although manual estimation of cell count on the slit-lamp exam has become the gold standard for grading ACI, several limitations exist~\cite{automated}. 
The SUN system is a subjective scoring and grading system that leads to marked differences by interobservers depending on different experiences.
Therefore, there have been efforts to quantitatively grade the ACI by the advances in optical imaging technology, such as Anterior Segment Optical Coherence Tomography (AS-OCT).
For example, some researchers~\cite{ius15, objective, Quantitative} calculate the number of hyperreflective dots representing AC cells and optical density ratio for flare qualification from AS-OCT images to diagnose ACI manually. However, manual methods are time-consuming and labour-intensive, so automated algorithms~\cite{automated, analysis, acc9} are developed to measure the number of cells seen in the AS-OCT images for ACI diagnosis.
For example, Baghdasaryan et al.~\cite{analysis} use the Image J Particle Analysis algorithm to analyze the degree of ACI automatically. Deep learning algorithms are also adopted to count and measure the size of AC cells~\cite{faq23}.
In summary, each modality data provides different advantages for the diagnosis of ACI in clinical, and none of the above methods pay attention to the integration of them.

\subsection{Fusion of Multimodal Data}
The growing accessibility of multimodal data has paved the way for the advancement of multimodal artificial intelligence solutions designed to comprehensively understand the intricacies of human health and disease~\cite{mbai}. Fusion of multimodal data is a crucial component of multimodal artificial intelligence solutions, offering the potential to address inference challenges with greater accuracy compared to unimodal approaches~\cite{mdl}.
In the field of medical artificial intelligence, the most common method of multimodal data fusion is to map the different modal data to a high-dimensional feature space, then fuse these features using different algorithms, and finally to achieve specific tasks, such as classification, segmentation or survival prediction~\cite{multisp, wcil, tmss, dmg}. The current methods usually adopt implicit modality interactions (i.e., self-attention and its variants).
For example, Zheng et al.~\cite{multisp} use a multimodal transformer for feature fusion and a decision-level fusion method to predict survival of nasopharyngeal carcinoma patients; Saeed et al.~\cite{tmss} adopt a transformer for fusion of different modality data to achieve the tasks of segmentation and survival prediction.
However, existing work neglects to inject explicit modality interactions, especially from expert knowledge and imaging properties.
As an illustration, ophthalmologists usually diagnose ACI by combining the advantages of slit-lamp and AS-OCT examination. Slit-lamp images reflect the overall condition of the eyeball and KPs intuitively, which are not easily observed on AS-OCT images.
However, the quantification of the number of AC cells and the degree of flare through slit-lamp may be largely influenced by the experience of ophthalmologists, while these indicators can be objectively and accurately quantified in AS-OCT images.
Therefore, based on implicit cross-modal interaction, we also perform cross-modal interaction based on these explicit features for diagnosing ACI.

\section{Methodology}
\subsection{Overview}
\begin{figure*}[t]
  \centering
  \includegraphics[width=1\linewidth]{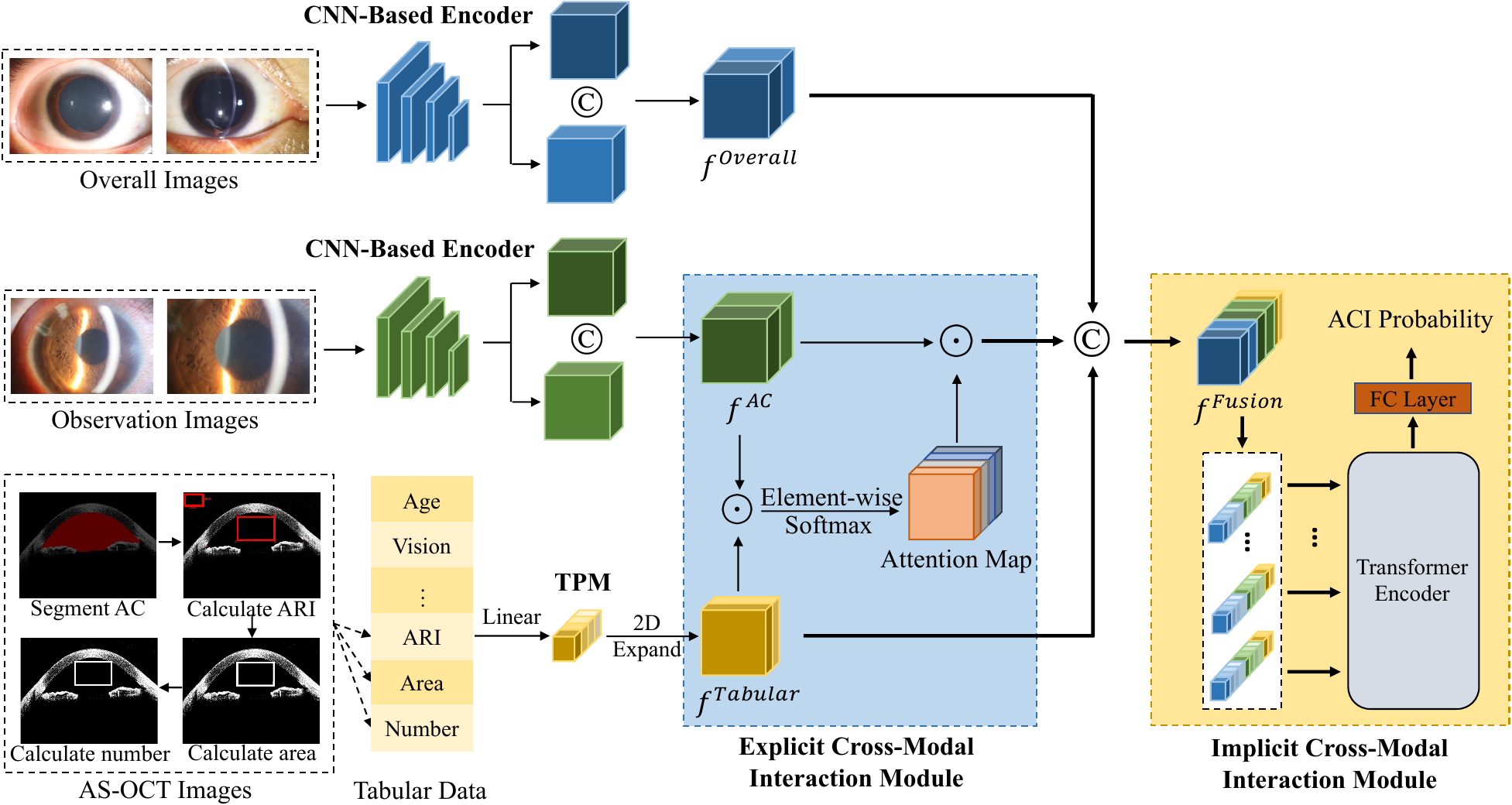}
  \caption{The overall architecture of our proposed \model. \textcircled{\small{C}} denotes the concatenate operation, while \textcircled{$\cdot$} denotes multiplying by element.}
  \label{fig:arc}
\end{figure*}

Our proposed \model exploits the advantages of each modality to enhance ACI diagnosis performance through explicitly and implicitly multimodal features fusion. As shown in Fig.~\ref{fig:arc}, \model consists of two Convolutional Neural Network-based Encoders (CNN-based Encoders), a Tabular Processing Module (TPM), an Explicit Cross-Modal Interaction Module (ECIM) and an Implicit Cross-Modal Interaction Module (ICIM).
To extract rich informativeness from slit-lamp images, we use two CNN-based encoders to encode two types of slit-lamp images (\ie the overall state images outside the eyeball and the observation images of AC), respectively. At the same time, the TPM first quantifies the features of AS-OCT images which are difficult to obtain from slit-lamp images, such as the number of AC cells, and then encodes these features together with other clinical data that may be closely related to ACI, such as best corrected vision acuity (BCVA), intraocular pressure and KPs to establish tabular feature as the explicit cross-modal interaction features.
Next, the ECIM generates attention maps by calculating the correlation between slit-lamp image feature maps and tabular feature maps, then integrating them into the slit-lamp feature maps, allowing the CNN-Based Encoder to focus on more effective informativeness of the slit-lamp images such as AC cells.
Afterwards, the ICIM employs a transformer-based network for implicit modality interactions.
Next, we depict the details of each module in the following subsections.

\subsection{Convolutional Neural Network-Based Encoder}
To obtain comprehensive information for the diagnosis of ACI, ophthalmologists usually use diffuse light to observe the overall state outside the eyeball and use slit light to observe the condition of different parts of the AC.
Therefore, based on this expert experience, we collect two types of slit-lamp images for each case, including two overall images that reflect the overall state outside the eyeball and two observation images that reflect various important indicators inside the AC.
Note that we bring two images in each type to magnify feature representation further because ophthalmologists often observe AC from different angles.
Technically, we can use any image encoder architecture to extract image features from each type. Here, we use a Resnet18~\cite{resnet18} as the backbone because it is a kind of classic and efficient network for feature extraction. Additionally, the complexity of Resnet18 matches our dataset, which can be verified in the later experiments.
Formally, in the process of overall image feature extraction, we take each image $I_i^{Overall} \in \mathbb{R}^{H \times W \times C},\,i=\{1,2\}$ as input, where $H$, $W$ and $C$ are the height, width, and channel respectively. The output image features from the encoder is represented as:
\begin{equation}
f_i^{Overall}=CNNEncoder(I_i^{Overall}), \, i=\{1,2\}.
\end{equation}
Then we take the concatenated results of two feature maps along the channel dimension $f^{Overall} \in \mathbb{R}^{H^{Overall} \times W^{Overall} \times C^{Overall}}$ as the final features. The process of the observation images is the same as above, and the final feature of AC observation images can be represented as $f^{AC}$.

\subsection{Tabular Processing Module}
The Tabular Processing Module (TPM) first quantifies the features of AS-OCT images, which are difficult to obtain from slit-lamp images, including area, number of AC cells and aqueous-to-air relative intensity (ARI).
To establish tabular feature maps, TPM then encodes these 3 features together with other 6 clinical features that may be closely related to ACI, including gender, age, BCVA, intraocular pressure, grading of keratic precipitates (KPs) and grading of aqueous flare.
All these 9 features are considered tabular data, which can be divided into numerical and categorical data.
Gender, grading of KPs, and grading of aqueous flare are categorical variables. In this context, we represent females with $0$ and males with $1$.
Age, BCVA, intraocular pressure, area and number of AC cells and aqueous-to-air relative intensity (ARI) are numerical.

In conclusion, the overall process is divided into two steps: quantification of AS-OCT image features and establishment of tabular feature maps.

\subsubsection{Quantification of AS-OCT Image Features}
First, we use U-net~\cite{unet} to segment the AC area and obtain the boundary between the AC and the cornea. We take the two endpoints of the boundary as the coordinate points of the AC angle. 
Next, a 2mm by 4mm (true intraocular distance) rectangular area $Area_1$ is specified in the centre of the AC, the long side of which is close to the boundary and parallel to the line connecting these two endpoints.
Then, a 1mm by 2mm rectangular area $Area_2$ is specified at the top left of the image, which calculates air intensity.
To calculate the number of AC cells, we perform threshold processing on the image. The points whose pixel value is higher than the threshold in $Area_1$ are candidate points for AC cells. The adjacent candidate points can be considered one AC cell. Finally, we calculate ARI according to the following formulas:
\begin{equation}
\begin{split}
ARI & = \frac{\frac{1}{n_{Area_1}-n_{CP}} \sum_{i \in {Area_1}-CP}PV_i}{\frac{1}{n_{Area_2}}\sum_{j \in {Area_2}}PV_j} \\
& = \frac{n_{Area_2} \sum_{i \in {Area_1}-CP}PV_i}{(n_{Area_1-n_{CP}})\sum_{j \in {Area_2}}PV_j},
\end{split}
\end{equation}
among which $n_{Area_1}$, $n_{Area_2}$ are the number of pixels in $Area_1$ and $Area_2$ respectively, $CP$ denotes all the candidate points in $Area_1$, $n_{CP}$ is the number of candidate points.
$PV_i$ and $PV_j$ denote the pixel value of the point $i$ and $j$ separately.

\subsubsection{Establishment of Tabular Feature Maps}
In the later experiments, we expand the dimensions of tabular data to make them consistent with that of image features for explicit and implicit cross-modal interactions. The specific method is as follows: we copy its value and construct a matrix matching the width and height of image features.
Then we obtain the tabular feature denoted as:
\begin{equation}
f^{Tabular}=TabularEncoder(Tabular\,data),
\end{equation}
where the height and width of the tabular feature are the same as $H^{Overall}$ and $W^{Overall}$, while the channel size is half of $C^{Overall}$. By ensuring uniformity of all values within each channel of $f^{Tabular}$, this operation enables it to serve as a channel attention mechanism that introduces feature maps from other modalities, thereby adaptively recalibrating channel-wise feature responses.

\subsection{Explicit Cross-Modal Interaction Module}
The ECIM introduces explicit modelling of the interdependencies among the feature map channels of the slit-lamp images by incorporating external expert knowledge (\ie $f^{Tabular}$), which consequently allows the CNN-based Encoder to concentrate on the more effective informativeness of the slit-lamp images, such as AC cells and KPs.

First, we multiply each element of $f^{Tabular}$ and $f_i^{AC},\,i=\{1,2\}$, then perform Softmax operations on the channel dimension to obtain the attention map. 
Then, we multiply each element of the attention map and $f^{AC}$ to obtain the attention-weighted feature map $f^{AC\_attention}$, allowing the CNN Based Encoder to focus on more detailed information of the slit-lamp images. The above process can be expressed as: 
\begin{equation}
Attention\,Map_i = \sigma(Mul(f^{Tabular},\,f_i^{AC}),\,i=\{1,2\},
\end{equation}
\begin{equation}
f_i^{AC\_attention} = Mul(Attention\,Map_i,\,f_i^{AC}),\,i=\{1,2\},
\end{equation}
where $\sigma$ and $Mul$ represent the element-wise Soft-max and multiplying by element operation. Note that, aligning with the unified attention mechanism with key, query, and value, $f^{Tabular}$ means query and $f_i^{AC}$ equals both key and value.

\subsection{Implicit Cross-Modal Interaction Module}
Self-attention-based transformer essentially aligns data from different modalities by calculating the similarity between inputs, making it very suitable for multimodal data fusion. Therefore, we propose ICIM, which is a transformer-based network for implicit modality interactions.

First, we concat $f^{Overall}$, $f^{AC\_attention}$ and $f^{Tabular}$ along the channel dimension to obtain the fusion feature $f^{Fusion}$. To reduce the computational effort, we transform the channel size of $f^{Fusion}$ to $C^{Fusion}$ (\ie from 2560 to 512 in our experiment). Since the Transformer encoder expects sequence-style as input, we treat each voxel of the feature along the channel dimension as a \textit{sequence of fusion feature}~\cite{transformer}, therefore the Transformer receives $(H^{Overall} \times W^{Overall})$ sequences as input and outputs a vector denoted as $V \in \mathbb{R}^{(H^{Overall} \times W^{Overall}) \times C^{Fusion}}$ to form  $C^{Fusion}$ predicted values as the final feature vector. Finally, we perform a fully connected (FC) layer and Softmax on the final feature vector to predict the probability of ACI. The prediction loss function can be expressed as:
\begin{equation}
Loss=-\frac{1}{B} \sum_{j=1}^B \sum_{i=1}^n y_{ji} log (\hat{y_{ji}}),
\end{equation}
where $B$ is the batch size, $n$ is the number of categories, $y_{ji}$ is the real probability that the $j$-th sample belongs to $i$-th category, and $\hat{y_{ji}}$ is the predicted probability that the $j$-th sample belongs to $i$-th category.

\section{Experiments}
\subsection{Subjects}
This study includes 104 eyes with uveitis and 35 healthy eyes of 82 participants from our collaborating hospital between May 23rd, 2018, and October 10th, 2022. All patients are diagnosed with unilateral or bilateral uveitis involving the anterior segment. During the same period, 35 healthy eyes without a history of ocular inflammation, injury, surgery, or other ocular diseases are recruited as a normal control group. Eyes with corneal opacity or shallow AC are excluded. Informed consent is obtained from each participant at enrollment. This study is approved by the Ethics Committee of State Key Laboratory of Ophthalmology, Zhongshan Ophthalmic Center, Sun Yat-sen University, Guangdong Provincial Key Laboratory of Ophthalmology and Visual Science, Guangdong Provincial Clinical Research Center for Ocular Diseases. Finally, we evenly divide the dataset into 5 parts and use 5-fold cross-validation to evaluate model performance.

\begin{table*}[htb]
\centering
    \caption{Description of Clinical Data.}
    \label{tab1}
    \begin{tabular}{cc|cc}
        \toprule
        Characteristics &  Number (Proportion) & Characteristics & Number (Proportion)\\
        \midrule
        Grading of aqueous flare & & Grading of ACI & \\
        0 & 55 (39.57$\%$) & 0 & 35 (25.18$\%$)\\
        0.5 & 3 (2.16$\%$) & 0.5 & 17 (12.23$\%$)\\
        1 & 36 (25.90$\%$) & 1 & 34 (24.46$\%$)\\
        2 & 28 (20.14$\%$) & 2 & 20 (14.39$\%$)\\
        3 & 17 (12.23$\%$) & 3 & 12 (8.63$\%$)\\
        4 & 0 (0$\%$) & 4 & 21 (15.11$\%$)\\
        Grading of KPs & & Gender (male) & 95 (68.35$\%$)\\
        \cline{3-4}
        0 & 41 (29.50$\%$) & Characteristics & Average± standard deviation\\
		\cline{3-4}                  
        1 & 24 (17.27$\%$) & Age & 38.11±14.84\\
        2 & 28 (20.14$\%$) & BCVA & 0.42±0.33\\                 
        3 & 44 (31.65$\%$) & Intraocular pressure & 12.38±4.74\\
        4 & 2 (1.44$\%$) & & \\
        \bottomrule
    \end{tabular}
\end{table*}

\subsection{Clinical Examination}
All participants undergo a comprehensive ocular examination by a uveitis specialist, including best-corrected vision acuity (BCVA), intraocular pressure measurements, slit-lamp observation of the AC, prefacing lens observation of the posterior segment and scoring of ACI according to SUN criteria~\cite{Standardization}. Clinical data collected includes gender, age, BCVA, intraocular pressure, keratic precipitates (KPs), aqueous flare and uveitis diagnosis. Eyes with uveitis are divided into two groups based on clinical evaluation: inactive eyes, when both AC cells and aqueous flare clinically are graded 0, and active eyes with any other grading score. In addition, all participants are examined by CASIA SS-1000 (Tomey, CASIA1), a Swept Source OCT (SS-OCT) which scans at a rate of 30,000 A-scans per second in the anterior segment scan (16mm*6mm) mode. Each eye contains 3 AS-OCT images close to 0°-180° cross-section.
The clinical data of participants are summarized in Table~\ref{tab1}.

\subsection{Experimental Settings and Metrics}
All models are implemented in PyTorch and trained on an RTX 3090 with 24GB memory. Specifically, we use stochastic gradient descent (SGD) with an initial learning rate $1e^{-3}$ to optimize the network. For a fair comparison, the epoch and the batch size are set to $100$ and $8$. 

We calculate the true positive (TP), true negative (TN), false positive (FP), and false negative (FN) values from the confusion matrix to obtain four widely-used evaluation metrics, including $accuracy$, $F_1\,score$, $precision$, and $recall$. We report the average indicators of the 5-fold cross-validation to evaluate the model performance.

\begin{table*}[htb]
\centering
\caption{Comparison With Other Methods Using a Single Modality Data.}
\begin{tabular}{cc|cccc}
\toprule
\multicolumn{2}{c|}{Method}                                              & \multicolumn{1}{l}{$Accuracy$} & $F_1\,score$       & $Precision$      & $Recall$         \\ \cmidrule{1-6}
\multicolumn{1}{c|}{\multirow{2}{*}{Overall images only}}     & Resnet18 & 0.878                        & 0.923          & 0.889          & 0.962          \\
\multicolumn{1}{c|}{}                                         & ViT      & 0.835                        & 0.897          & 0.854          & 0.952          \\ \cmidrule{1-6}
\multicolumn{1}{c|}{\multirow{2}{*}{Observation images only}} & Resnet18 & 0.886                        & 0.924          & 0.908          & 0.943          \\
\multicolumn{1}{c|}{}                                         & ViT      & 0.834                        & 0.897          & 0.836          & 0.971          \\ \cmidrule{1-6}
\multicolumn{1}{c|}{\multirow{2}{*}{AS-OCT images only}}      & Resnet18 & 0.921                        & 0.950          & 0.921          & 0.981          \\
\multicolumn{1}{c|}{}                                         & ViT      & 0.784                        & 0.873          & 0.791          & 0.981          \\ \cmidrule{1-6}
\multicolumn{1}{c|}{\multirow{2}{*}{Clinical data only}}      & MLP      & 0.843                        & 0.908          & 0.837          & \textbf{1.000} \\
\multicolumn{1}{c|}{}                                         & SVM      & 0.906                        & 0.941          & 0.914          & 0.971          \\ \cmidrule{1-6}
\multicolumn{1}{c|}{Multimodal data}                          & EiCI-Net & \textbf{0.935}               & \textbf{0.958} & \textbf{0.950} & 0.971          \\ \bottomrule
\end{tabular}
\label{tab2}
\end{table*}

\subsection{Comparison With Other Methods Using Single-modal Data}
To prove the superiority of our proposed \model, we compare the performance with four classification methods by using a single modality data, including:

(1) Resnet18~\cite{resnet18}: a kind of classical CNN using overall state images, slit-lamp observation images or AS-OCT images as input only;

(2) ViT~\cite{vit}: a kind of network of self-attention-based architecture using overall state images, observation images, or AS-OCT images only;

(3) Multilayer Perceptron (MLP): a network of three hidden layers for classification using clinical data only;

(4) Support Vector Machine (SVM)~\cite{svm}: a machine learning method for classification using clinical data only.

\begin{table*}[htb]
\centering
\caption{Comparison With Other Methods Using Multimodal Data.}
\begin{tabular}{ccccc}
\toprule
Method        & $Accuracy$       & $F_1\,score$     & $Precision$       & $Recall$ \\ \cmidrule{1-5}
Multi-transSP & 0.921            & 0.950            & 0.912             & 0.990  \\
TMSS          & 0.907            & 0.944            & 0.897             & \textbf{1.000}  \\
\model        & \textbf{0.935}   & \textbf{0.958}   & \textbf{0.950}    & 0.971  \\ \bottomrule
\end{tabular}
\label{tab3}
\end{table*}

\begin{table*}[htb]
\centering
\caption{Ablation Study of the Contribution of Key Components in \model.}
\begin{tabular}{ccccc|cccc}
\toprule
\multicolumn{5}{c|}{Key Component}           & \multirow{2}{*}{$Accuracy$} & \multirow{2}{*}{$F_1\,score$} & \multirow{2}{*}{$Precision$} & \multirow{2}{*}{$Recall$} \\
$Encoder^{Overall}$ & $Encoder^{AC}$ & TPM & ECIM & ICIM &                           &                           &                            &                         \\ \cmidrule{1-9}
\checkmark       & \checkmark       & \checkmark   &      & \checkmark    & 0.893                     & 0.932                     & 0.888                      & 0.981                   \\
\checkmark       & \checkmark       & \checkmark   & \checkmark    &      & 0.893                     & 0.936                     & 0.893                      & \textbf{0.990}          \\
        & \checkmark       & \checkmark   & \checkmark    & \checkmark    & 0.928                     & 0.954                     & 0.937                      & 0.971                   \\
\checkmark       &         & \checkmark   &      & \checkmark    & 0.863                     & 0.913                     & 0.888                      & 0.942                   \\
\checkmark       & \checkmark       &     &      & \checkmark    & 0.886                     & 0.929                     & 0.885                      & 0.981                   \\
\checkmark       & \checkmark       & \checkmark   & \checkmark    & \checkmark    & \textbf{0.935}            & \textbf{0.958}            & \textbf{0.950}             & 0.971                   \\ \bottomrule
\end{tabular}
\label{tab4}
\end{table*}

The experimental results are shown in Table~\ref{tab2} on which we have several observations:
(1) Our method has obtained the consistent best performance under the metrics of $accuracy$, $F_1\,score$, and $precision$, and competitive performance under $recall$, which is superior to other methods based on single modality data. This indicates that fusing multimodal data to diagnose ACI is effective.
(2) The performance of the Resnet18 using slit-lamp images only is almost superior to that of ViT. This may be due to the high complexity of ViT and the small size of our dataset, which results in over-fitting, whereas the complexity of Resnet18 matches our dataset to achieve better performance.
(3) Resnet18 trained by AS-OCT images performs second only to \model, outperforming slit-lamp image-based methods. This may be because the background of AS-OCT images is relatively simple, which facilitates the automatic extraction of features, while there are several parts in the slit-lamp images that interfere with the diagnosis, such as eyelashes, slit-lamp bands, etc.
(4) The performance of SVM based on clinical data is superior to the methods based on slit-lamp images. From the point of view of the modality itself, the clinical data is relatively streamlined and contains indicators that are strongly correlated with ACI, such as aqueous flare, KPs, etc.
While slit-lamp image modality is relatively redundant, and the information it provides for diagnosing tasks largely depends on the feature extraction method. In a word, traditional feature extraction methods cannot extract the features of slit-lamp images relatively effectively.

\subsection{Comparison With Other Methods Using Multi-modal Data}
To prove the superiority of our proposed \model, we compared the performance with the other two fusion methods~\cite{multisp, tmss} in the field of medical artificial intelligence.
Multimodal data includes two types of slit-lamp images: AS-OCT images and clinical data.
Two fusion methods are as follows:

(1) Multi-transSP~\cite{multisp}: implicit feature-level fusion and decision-level fusion methods are used to predict the survival of nasopharyngeal carcinoma patients;

(2) TMSS~\cite{tmss}: a transformer-based network is used for implicit fusion of different modality data to achieve the tasks of segmentation and survival prediction.

Because our task differs from the above methods, only the feature fusion part of their methods was used in the comparison experiment.

The experimental results are shown in Table~\ref{tab3} on which we have several observations:
(1) The performance of our method is better than the two methods.
Specifically, the $accuracy$ of our model is 0.014 and 0.028 higher than that of Multi-transSP~\cite{multisp} and TMSS~\cite{tmss}, respectively.
While the $F_1\,score$ is 0.008 and 0.014 higher than that of Multi-transSP~\cite{multisp} and TMSS~\cite{tmss}, respectively.
The results demonstrate that the method using jointly implicit and explicit cross-modal interactions is superior to the method using only implicit cross-modal interactions;
(2) Our method achieves relatively low recall compared to the other two methods, but both the precision and recall of our method exceed 0.950, which already meets the standards for clinical diagnosis;
(3) Several methods using single modality data (\ie Resnet18 using AS-OCT images only and SVM using clinical data only) achieve performance comparable to the above methods, which may be because multimodal data contains a large amount of redundant information, making it difficult to interact and fuse effective information using implicit cross-modal interaction method only, thereby limiting the performance of the model.
Our model explicitly applies this information directly to the interaction between modalities, thus achieving higher performance.

\subsection{Ablation Study}
Finally, we conduct an ablation study to investigate the contribution of key components in our \model. The experimental setup can be summarized as follows:

(1) \model without ECIM;

(2) \model without ICIM but a convolutional layer for feature fusion;

(3) \model without CNN-Based Encoder for feature extraction of the overall state images;

(4) \model without CNN-Based Encoder for feature extraction of the observation images;

(5) \model without TPM.

We should note that the ECIM generates cross-modal attention maps based on $f^{AC}$ and $f^{Tabular}$. Therefore, when the CNN-Based Encoder for feature extraction of the observation images or TPM is removed in the ablation experiments, the ECIM cannot work.

\begin{figure*}[t]
    \centering
    \includegraphics[width=0.8\linewidth]{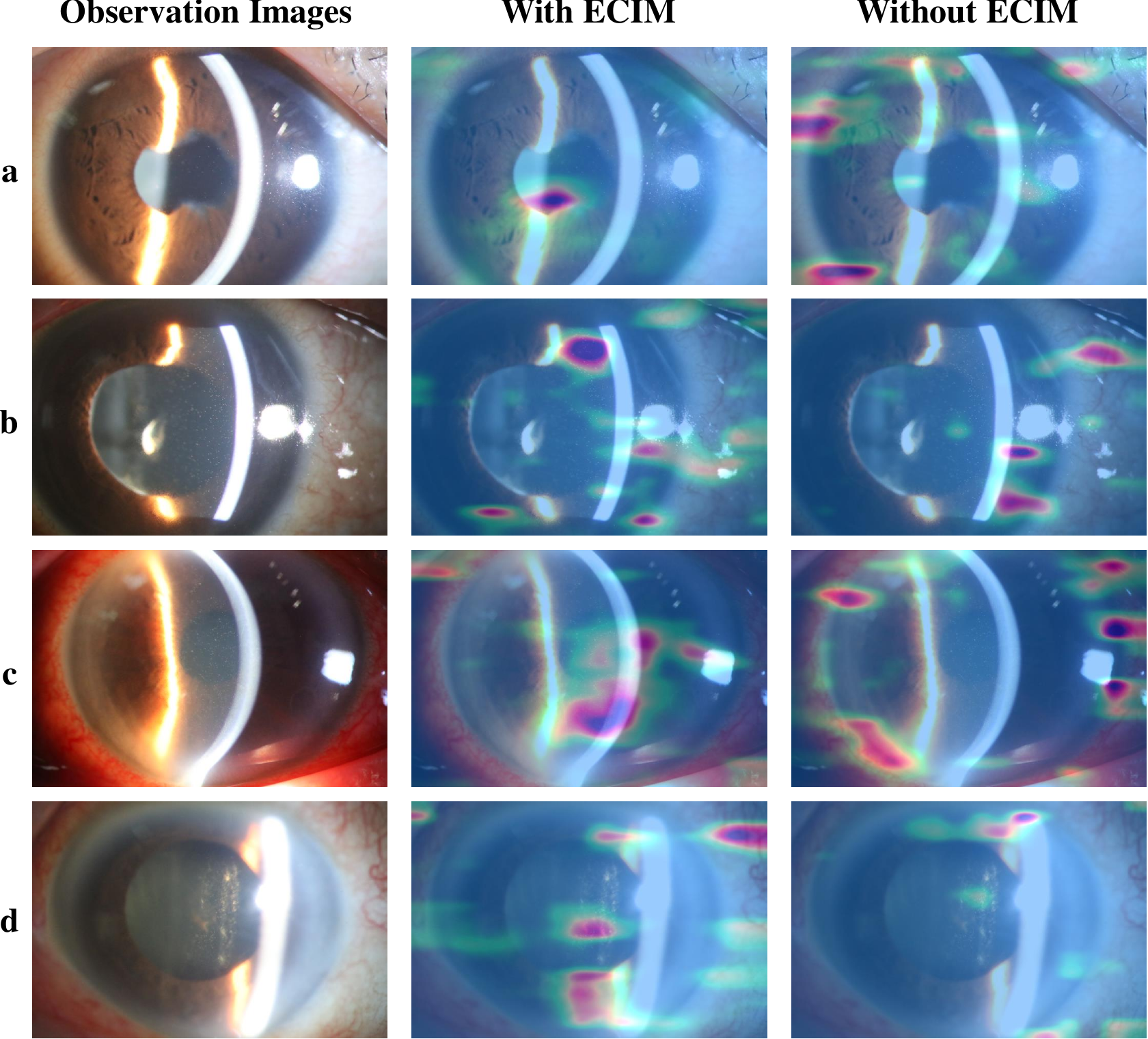}
    \caption{Visualization of \model w or w/o ECIM. a. the pupil is not round; b$\sim$c. AC cells; d. the vitreous body and lens are turbid.}
    \label{fig:vis}
\end{figure*}

The quantitative results are shown in Table~\ref{tab4} on which we have several observations:
(1) When we remove any module, the model's performance decreases to varying degrees, indicating that each module is valid.
(2) When $Encoder^{AC}$ or TPM is removed, the most significant drop in model performance occurs ($accuracy$ decreased by 0.072 and 0.049 respectively, while $F_1\,score$ decreased by 0.045 and 0.029 respectively), indicating that observation images and tabular data are more important modalities. This is consistent with the results shown in Table~\ref{tab2}, where observation images or clinical data-based methods perform better than other modality-based methods.
(3) When $Encoder^{Overall}$ is removed, the model performance degradation is least noticeable, indicating that overall image modality is secondary. This is consistent with the results shown in Table~\ref{tab2}, where overall image-based methods perform less well than other modality-based methods.
(4) When ECIM or ICIM is removed, the degree of decrease in $accuracy$ is the same at 0.042, which demonstrates that the explicit cross-modal interaction and the implicit cross-modal interaction are almost equally important.

In addition, we employ Grad-CAM~\cite{grad} to visually observe the area of interest of \model, proving that ECIM allows the CNN-Based Encoder to focus on more effective informativeness of the slit-lamp images, as shown in Fig.~\ref{fig:vis}.
The first column images in Fig.~\ref{fig:vis} are the original observation images, the second column shows heat maps generated based on $f^{AC\_attention}$ with ECIM and the third ones are heat maps generated based on $f^{AC}$ without ECIM. 
The specific process of heat map generation is that we use the gradients of logits for ACI, flowing into the final convolutional layer of CNN-based Encoder, which is used for feature extraction of observation images to produce the heat map highlighting the important regions in the image. 
In the heat maps, the red area is what the model focuses on.

In Fig.~\ref{fig:vis}, we can see that the model with ECIM focuses more on indicators that ophthalmologists will pay attention to in clinical examinations:
(1) the pupil in Fig.~\ref{fig:vis}a is not round which reflects uveitis in the past;
(2) AC cells in Fig.~\ref{fig:vis}b and Fig.~\ref{fig:vis}c which is one of the most important indicators of ACI;
(3) the vitreous body and lens in Fig.~\ref{fig:vis}d are turbid, which suggests the possibility of ACI.
In contrast, the model without ECIM pays less attention to this information, which proves the effectiveness of the ECIM in the cross-modal interaction.

\section{Limitations}
Firstly, our proposed method cannot predict the grade of inflammation. It is difficult to accurately grade the ACI clinically, especially for a grade larger than 4, because the AC cells of patients with high-grade inflammation are usually very dense under slit-lamp observation and may appear cloudy on AS-OCT images, which is difficult to evaluate quantitatively by naked eyes or automated algorithms. Therefore, our subsequent research will focus on the grade classification of ACI.

Secondly, due to the lack of publicly available multi-modal datasets for ACID, we can only verify the effectiveness of our method on a relatively limited number of in-house datasets. To more comprehensively evaluate the effectiveness of our method and explore more effective ACID methods, we plan to collect multi-centre and large-scale datasets, which are already in progress.

\section{Conclusions}
In this paper, we proposed a cross-modal interaction network called \model to jointly learn the explicit and implicit feature correlations among all modalities to diagnose ACI. 
The experimental results on our in-house dataset demonstrated that it is superior to the state-of-the-art classification methods.
Although the diagnosis of ACI based on multi-modal data is more accurate, it is difficult to collect all modality data in clinical practice, which limits the in-depth research on diagnosing ACI using multi-modal data. We collect and organize a multi-modal dataset and plan to make it publicly available.
We hope that our contribution will draw more attention from the community, fostering a more expansive perspective on ACID using multi-modal data.

\section*{Acknowledgment}

This research was partially supported by the National Natural Science Foundation of China under grant No. 62106218, and the Key Research and Development Program of Zhejiang Province under grant No. 2024C01104.

\bibliographystyle{IEEEtran}
\bibliography{IEEEabrv,main}

\end{document}